\documentclass[10pt,twocolumn,letterpaper]{article}

\usepackage{wacv}
\usepackage{times}
\usepackage{epsfig}
\usepackage{graphicx}
\usepackage{amsmath}
\usepackage{amssymb}

\usepackage{multirow}
\usepackage{booktabs}

\usepackage{caption}
\usepackage{subcaption}

\usepackage[pagebackref=true,breaklinks=true,letterpaper=true,colorlinks,bookmarks=false]{hyperref}
\makeatletter
\@namedef{ver@everyshi.sty}{}
\makeatother
\usepackage{tikz,pgfplots}
\usetikzlibrary{plotmarks,shapes,arrows,positioning,3d}
\usetikzlibrary{arrows.meta}
\usetikzlibrary{decorations.text}
\pgfplotsset{compat=newest} 

\usepackage{bm}

\newcommand{\R}{\mathbb{R}}

\usepackage{color}

\newlength\figureheight
\newlength\figurewidth

\renewcommand{\paragraph}[1]{\medskip\noindent{\bf#1}\ \ }

\def\wacvPaperID{1073}

\wacvfinalcopy

\ifwacvfinal

\fi
\ifwacvfinal
\usepackage[breaklinks=true,bookmarks=false]{hyperref}
\else
\usepackage[pagebackref=true,breaklinks=true,colorlinks,bookmarks=false]{hyperref}
\fi

\usepackage[capitalize,nameinlink]{cleveref}
\crefname{section}{Sec.}{Sects.}
\crefname{proposition}{Prop.}{Props.}
\crefname{lemma}{Lem.}{Lems.}
\crefname{model}{Mod.}{Mods.}
\crefname{appendix}{App.}{Apps.}    

\begin{document}

\title{Novel View Synthesis via Depth-guided Skip Connections}

\author{Yuxin Hou \qquad Arno Solin \qquad Juho Kannala\\
Department of Computer Science, Aalto University, Espoo, Finland\\
{\tt\small firstname.lastname@aalto.fi}
}

\maketitle

\begin{abstract}
We introduce a principled approach for synthesizing new views of a scene given a single source image. Previous methods for novel view synthesis can be divided into image-based rendering methods (\eg, flow prediction) or pixel generation methods. Flow predictions enable the target view to re-use pixels directly, but can easily lead to distorted results. Directly regressing pixels can produce structurally consistent results but generally suffer from the lack of low-level details. In this paper, we utilize an encoder--decoder architecture to regress pixels of a target view. In order to maintain details, we couple the decoder aligned feature maps with skip connections, where the alignment is guided by predicted depth map of the target view. Our experimental results show that our method does not suffer from distortions and successfully preserves texture details with aligned skip connections.
\end{abstract}

\section{Introduction}

\begin{figure}[!t]
  \centering
  \setlength{\figurewidth}{.2\columnwidth}
  \setlength{\figureheight}{\figurewidth}

  \newcommand{\figg}[1]{\includegraphics[width=.98\figurewidth]{./fig/fig1/#1}}

  \newcommand{\figrow}[2]{%
     \node [draw=white,thick,minimum width=\figurewidth,inner sep=0] at
       ({0*\figurewidth},#2) {\figg{#1_source}};%
     \node [draw=white,thick,minimum width=\figurewidth,inner sep=0] at
       ({1*\figurewidth},#2) {\figg{#1_gt}};%
     \node [draw=white,thick,minimum width=\figurewidth,inner sep=0] at
       ({2*\figurewidth},#2) {\figg{#1_flow}};%
     \node [draw=white,thick,minimum width=\figurewidth,inner sep=0] at
       ({3*\figurewidth},#2) {\figg{#1_gen}};%
     \node [draw=white,thick,minimum width=\figurewidth,inner sep=0] at
       ({4*\figurewidth},#2) {\figg{#1_ours}};%
  }

  \begin{tikzpicture}

  \def\myarray{{"Source view","target view","image-based rendering","pixel generation","our method"}}
  \foreach \i in {0,1,2,3,4}
     \node[text width=.9\figurewidth,align=center,text centered, text depth = 0cm, execute at begin node=\setlength{\baselineskip}{1.8ex}] at ({\figurewidth*\i},{-.65*\figureheight}) {\scriptsize \sc \vphantom{$^\dagger$}\pgfmathparse{\myarray[\i]}\pgfmathresult};

  \figrow{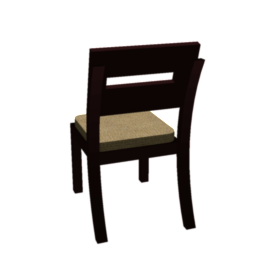}{0\figureheight}  

  \figrow{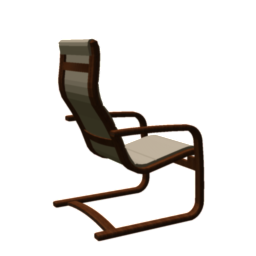}{1.0\figureheight}
  \figrow{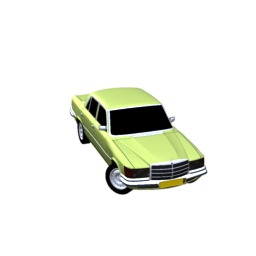}{2.0\figureheight}
\figrow{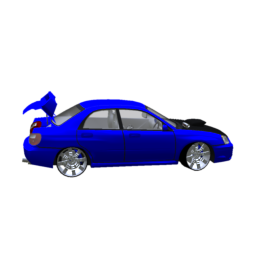}{2.8\figureheight}
        
  \end{tikzpicture}   

  \caption{Results of image-based rendering methods suffer from distortion, while the results of direct pixel generation methods lack detailed features. Our method has the benefits from both warping methods and pixel generation methods.}
  \label{fig:fig1}
\end{figure}
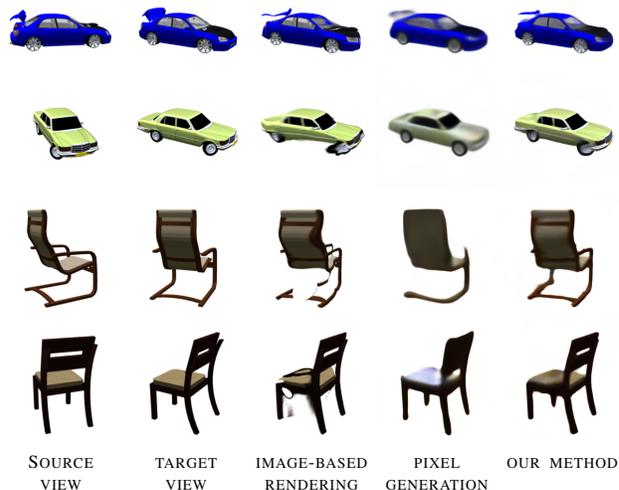

Novel view synthesis (NVS) is the task of generating new images of a scene given single or multiple inputs of the same scene (see, \eg, in \cref{fig:fig1}: given one image of the object, we generate a new image of the object from a novel viewpoint). NVS has various applications. For example, it can be used in virtual reality applications, where capturing all possible viewpoints of real-world scenes is impractical. With NVS one can just capture few images to offer a seamless experience.  Moreover, NVS enables users to edit images more freely (\eg, rotating products interactively in 3D for online shopping). 

Generally, to solve the NVS task, comprehensive 3D understanding of the scene by the model is important. Given the 3D geometry, we can render target views with 3D model-based rendering techniques. In that case, some methods estimate the underlying geometry of the scene with 3D representations like voxels \cite{choy20163d}, and mesh \cite{kato2018neural}, but these methods can be computationally expensive. Unlike traditional 3D model-based rendering, image-based rendering (IBR) methods render novel views directly from input images. Some IBR methods predict the appearance flow directly without geometry~\cite{ji2017deep, zhou2016view}. Some IBR methods render with explicit geometry, such as 3D warping with depth maps, which use a geometric transformation to obtain pixel-to-pixel correspondences~\cite{chen2019monocular}. Since the pixels from input views can be re-projected to the target view directly, original low-level details of the scene like colours and textures are well-preserved. However, estimating the accurate correspondences can be challenging, especially for single input views in texture-less regions and occlusion regions, and the failures can easily lead to distorted synthesized results. On the other hand, some methods attempt to regress pixels directly~\cite{tatarchenko2016multi, yang2015weakly}. These pixel generation methods can generate structurally consistent geometric shapes, but the visual quality of generated results are worse than methods that exploit correspondence since the lack of detailed features. \cref{fig:fig1} shows the limited performance of pixel generation methods and image-based rendering methods.

\begin{figure*}[!t]
\centering
   \includegraphics[width=1.95\columnwidth]{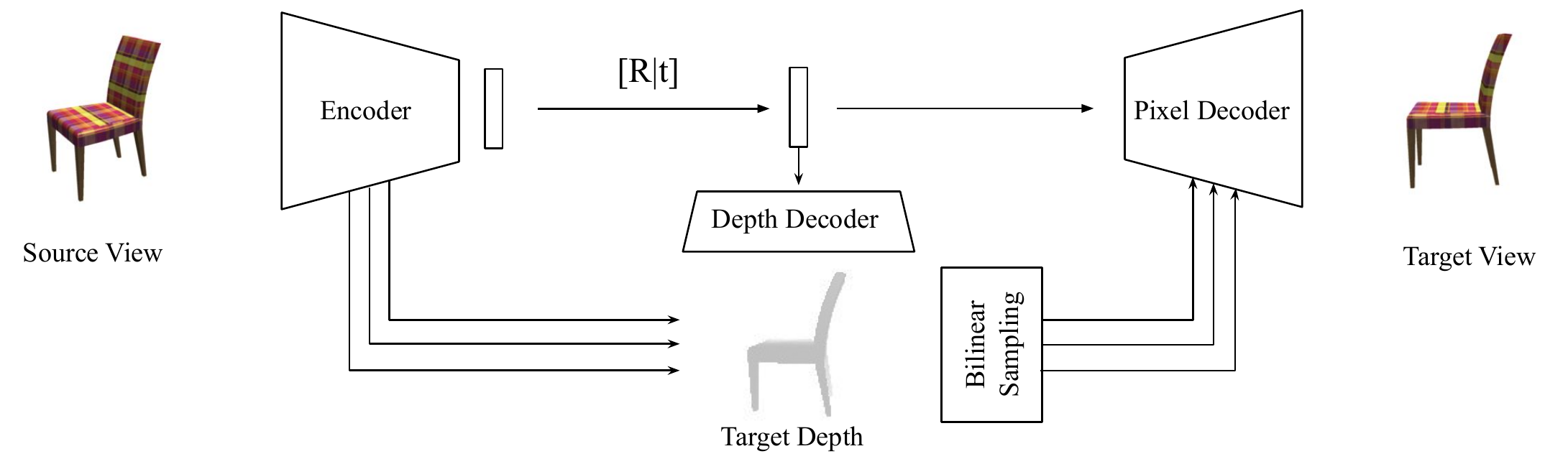}
   \caption{Overview of our architecture. There are three main components: the encoder, the depth decoder, and the pixel decoder. We apply the given geometric transformation on the latent code from the source view directly to obtain the latent code for the target view. Given the transformed latent code, we predict the depth map of the target view, and use the predicted depth to warp feature maps of source view to the target with bilinear sampling. The warped feature maps are then passed to the decoder as skip connections to assist pixel generation.}
\label{fig:arch}
\end{figure*}

In this paper, we aim to combine the advantages of both image-based rendering methods and direct pixel generation methods. The main benefit of image-based rendering methods is that they can exploit the correspondence pixels of target views in the source view and re-use pixels, so the predictions will not lose detailed features. Similarly, we decide to utilize the skip connections between the encoder and the decoder to transfer low-level features. Skip connections with U-net architectures~\cite{ronneberger2015u} have proved to be useful for passing feature information to lower layers in vision tasks like semantic segmentation, where the output and the input are well-aligned spatially. However, for novel view synthesis, the skip connections cannot be applied immediately because of the different shapes of the input view and the target view. To address the problem, we predict the depth map of the target map, warping multi-level feature maps before passing them to the generation decoder. Compared to warp image pixels directly, the generation decoder can exploit learned prior knowledge to `correct' distortion.

In conclusion, we propose an image generation pipeline with transforming auto-encoders which utilize warped skip connections to transfer low-level features. We compare our model against both state-of-the-art pixel generation methods and image-based rendering methods, demonstrating that our method can alleviate common issues like distortion and lack of details.

\section{Related Work}

\paragraph{Geometry-based view synthesis} A large body of research attempts to solve the novel view synthesis problem via explicitly modelling the underlying 3D geometry. To represent the 3D structure, common 3D representations like voxel~\cite{choy20163d,nguyen2019hologan, tulsiani2017multi, olszewski2019transformable, kar2017learning}, point cloud~\cite{wiles2020synsin}, mesh~\cite{kato2018neural}, and layered representations \cite{shade1998layered, tucker2020single, zhou2018stereo} are widely applied. Meshes and point-clouds can suffer from sparsity. Voxel-based methods are limited with the type of scenes and resolutions because of the memory constraints. Layered representations rely on a large number of layers to reach good quality. Apart from discrete representations, recent interests in continuous 3D-structure-aware representation also show promising results~\cite{oechsle2019texture, sitzmann2019scene}. In this paper, though we do not use explicit 3D representation, we predict the depth map from single view (2.5D) to help understand the 3D geometry of the input view and the depth map will be used to guide aligned skip connections. 

\paragraph{Image generation with disentangled representation} Many deep generative networks are capable of generating photorealistic images, but to solve the novel view synthesis task, networks generally required explicitly disentangled representations, like decoupling the pose and identity features. Cheung~\etal~\cite{cheung2014discovering} utilize an auto-encoder to learn disentangled latent units representing various factors of variations (poses, illumination conditions, \etc) and generate novel manipulations of images. Tatarchenko~\etal~\cite{tatarchenko2016multi} use separate convolutional layers and fully connected layers to process the image and the angle independently, then the separate latent representations are merged before up-convolutional layers. Casale~\etal\cite{casale2018gaussian} divide the feature vector into an object feature and a view feature, and utilize GP priors to model correlations between views. Inspired by traditional graphics rendering pipelines, Zhu~\etal~\cite{zhu2018visual} build the 3D representation from three independent factors: shape, viewpoint, and texture. Besides factorizing latent representations, some methods use equivariant representations to handle transformations. Hinton~\etal~\cite{hinton2011transforming} proposed Transforming auto-encoders (TAE) to model both 2D and 3D transformations of simple objects. Generally, directly generated images may suffer from blurriness, lack of texture details, or inconsistency of identity. 

\paragraph{Image-based Rendering} Image-based rendering re-uses the pixels from source images to generate target views. Previous image-based rendering methods can be classified into three categories according to how much geometric information is used: rendering without geometry, rendering with implicit geometry, and rendering with explicit geometry (approximate or accurate geometry) \cite{shum2000review}.  Some traditional methods construct a continuous representation of the plenoptic function from observed discrete samples with unknown scene geometry, like building lightfield~\cite{levoy1996light} or lumigraph~\cite{gortler1996lumigraph, buehler2001unstructured} with dense input views. Recently, some learning-based methods predict correspondence without geometry directly. Zhou~\etal~\cite{zhou2016view} use separate encoders for input images and viewpoint transformation and predict appearance flow filed directly. Ji~\etal~\cite{ji2017deep} used a rectification network before generating dense correspondences between two input views, and the view morphing network finally synthesizes the target middle view via sampling and blending. When the depth information is available, 3D warping can be used to render nearby viewpoints.  Some methods estimate depth maps from multi-view inputs \cite{debevec1996modeling, flynn2016deepstereo}. Choi~\etal~\cite{choi2019extreme} estimate a depth probability volume that accumulated from multi inputs rather than a single depth map of the novel view. Chen~\etal~\cite{chen2019monocular} use the TAE to predict the depth map of the target views directly. Our method does not re-use the pixels from inputs directly; instead, we re-use the feature maps extracted from the input view.

To combine the advantages of image-based rendering and image generation, Park~\etal~\cite{park2017transformation} used two consecutive encoder--decoder networks, first predicting a disocclusion-aware flow and then refining the transformed image with a completion network. Sun~\etal~\cite{sun2018multi} proposed a framework to aggregate flow-based predictions from multiple input views and the pixel generation prediction via confidence maps. In this paper, we present a different way that can bring the power of explicit correspondence to image generation via skip connections.

\section{Methods}
Our architecture consists of three main parts: the encoder $\phi$, the depth prediction decoder $\psi_\mathrm{d}$, and the pixel generation decoder $\psi_\mathrm{p}$. \cref{fig:arch} shows the overview of our pipeline. The encoder extracts feature maps and latent representation for the source view firstly. To exploit geometric transformation explicitly, we apply the given transformation matrix on the latent representation from the source view to obtain the latent representation of the target view, which will be passed to the depth decoder and the pixel decoder. To take advantage of correspondence pixels in the source view, we predict the depth map for the target view given the transformed latent code. Then we can use the estimated depth map to find dense correspondences between target and source views. Instead of warping the source image into the target view, we warp the multi-level feature maps extracted from the encoder via bilinear sampling and then pass them to the decoder as skip connections, transferring low-level details to assist final pixel regression.

\subsection{Transformable Latent Code}
Inspired by \cite{chen2019monocular} that applying the transformation matrix on latent code directly to predict depth map for target view, we also adopt the idea of using a TAE to learn a compact latent representation that are transformation equivariant. Given the source image $I_s$, the learnt latent code $z_\mathrm{s} = \phi(I_{\mathrm{s}})$ can be regarded as a set of points $z_\mathrm{s}\in \R^{n\times3}$ extracted by encoder $\phi$. Then the representation is multiplied with the given transformation $T_{\mathrm{s}\to \mathrm{t}} = [R\,|\,t]_{\mathrm{s}\to \mathrm{t}}$ to get the transformed latent code for the target view:
\begin{equation}
\tilde{z}_\mathrm{t} = T_{\mathrm{s}\to \mathrm{t}}\cdot \dot{z}_\mathrm{s},
\end{equation}
where $\dot{z}_\mathrm{s}$ is the homogeneous representation of $z_\mathrm{s}$. Intuitively, training in this way will encourage the latent code to encode 3D position information for features.

\subsection{Depth-guided Skip Connections}
Since \cite{chen2019monocular} is an image-based rendering method, the quality of prediction results relies on the accuracy of estimated depth maps. However, the monocular depth estimation with the TAE architecture (w/o skip connections) is challenging. In that case, pure image-based rendering methods can lead to distortion easily because of the unstable depth prediction. Also with a monocular input, image-based rendering cannot inpainting the missing parts since they do not have correspondences in source views. In this work, to alleviate the mentioned limitations, we decide to synthesis the target view with a pixel generation pipeline, regressing the pixel value with the pixel decoder $\psi_\mathrm{p}$.

Rethinking about the TAE architecture used in \cite{chen2019monocular} and the design of the equivariant latent code $z$, though $z\in \R^{n\times3}$ might be sufficient for encoding position predictions for features and then be mapped into depth maps, regressing the pixels directly can be difficult. It is mainly because the downsample and upsample process can lose much detailed information, especially for the view that includes many small objects or rich textures. Generally, the skip connections have proved effective in recovering fine-grained details, but it cannot be used directly with the TAE architecture since the shape of output changed. In that case, there are two decoders in our framework, one for depth prediction $\psi_\mathrm{d}$ and one for pixel generation $\psi_\mathrm{p}$. After getting the depth estimation for target view $\tilde{D_\mathrm{t}} = \psi_\mathrm{d}(\tilde{z})$, we use the depth map to warp the feature maps $F^{i}$ at different level $i$. In order to maintain texture details, we also have a feature map that maintains image resolution, which we call \textit{conv0}. Given the camera intrinsic matrix $K$, the relative pose $T_{\mathrm{t}\to \mathrm{s}}$ and the predicted depth map of the target view $\tilde{D_\mathrm{t}}$, we can find the correspondences in the source view in the following way \cite{zhou2017unsupervised}:
\begin{equation}
p_\mathrm{s} \sim K\,T_{\mathrm{t}\to \mathrm{s}}\,\tilde{D}_\mathrm{t}(p_\mathrm{t})\,K^{-1}\,p_\mathrm{t},
\end{equation}
where $p_\mathrm{t}$ and $p_\mathrm{s}$ denote the homogeneous coordinates of a pixel in the target view and source view respectively. Since the obtained correspondences $p_s$ are continuous values, we use differentiable bilinear sampling that interpolates the values of the 4-pixel neighbours of $p_\mathrm{s}$ to approximate $F_\mathrm{i}(p_\mathrm{s})$. The warped feature maps can be represented as $\tilde{F}_\mathrm{t}^\mathrm{i}$, which will be passed to the pixel decoder $\psi_\mathrm{p}$ for concatenation.

Guided by predicted depth maps, the skip connections of warped feature maps enable the method to benefit from establishing explicit correspondences and maintain the low-level details. Also compared to image-based rendering that warp pixels directly, using multi-level skip-connections of warped feature maps helps to exploit learned prior information and avoid the loss of information.

\begin{figure*}[!t]
  \centering
  \setlength{\figurewidth}{.166\columnwidth}
  \setlength{\figureheight}{\figurewidth}

\begin{subfigure}[b]{.48\textwidth} 

  \newcommand{\figg}[1]{\includegraphics[width=.98\figurewidth]{./fig/chair/#1}}

  \newcommand{\figrow}[2]{%
     \node [draw=white,thick,minimum width=\figurewidth,inner sep=0] at
       ({0*\figurewidth},#2) {\figg{#1_source}};%
     \node [draw=white,thick,minimum width=\figurewidth,inner sep=0] at
       ({1*\figurewidth},#2) {\figg{#1_gt}};%
     \node [draw=white,thick,minimum width=\figurewidth,inner sep=0] at
       ({2*\figurewidth},#2) {\figg{#1_mv3d}};%
     \node [draw=white,thick,minimum width=\figurewidth,inner sep=0] at
       ({3*\figurewidth},#2) {\figg{#1_sun}};%
     \node [draw=white,thick,minimum width=\figurewidth,inner sep=0] at
       ({4*\figurewidth},#2) {\figg{#1_chen}};%
        \node[draw=white,thick,minimum width=\figurewidth,inner sep=0] at
       ({5*\figurewidth},#2) {\figg{#1_ours}};
  }

  \begin{tikzpicture}

 \node[text width=.9\figurewidth,align=center,text centered,text depth = 0cm] at ({\figurewidth*0},{-.65*\figureheight}) {\footnotesize Input};
 \node[text width=.9\figurewidth,align=center,text centered,text depth = 0cm] at ({\figurewidth*1},{-.65*\figureheight}) {\footnotesize Target};
 \node[text width=.9\figurewidth,align=center,text centered,text depth = 0cm] at ({\figurewidth*2},{-.65*\figureheight}) {\footnotesize \cite{tatarchenko2016multi}};
 \node[text width=.9\figurewidth,align=center,text centered,text depth = 0cm] at ({\figurewidth*3},{-.65*\figureheight}) {\footnotesize \cite{sun2018multi}};
 \node[text width=.9\figurewidth,align=center,text centered,text depth = 0cm] at ({\figurewidth*4},{-.65*\figureheight}) {\footnotesize \cite{chen2019monocular}};
 \node[text width=.9\figurewidth,align=center,text centered,text depth = 0cm] at ({\figurewidth*5},{-.65*\figureheight}) {\footnotesize Ours};

  \figrow{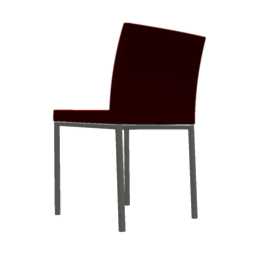}{0}  
  \figrow{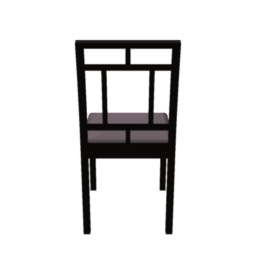}{1.0\figureheight}
  \figrow{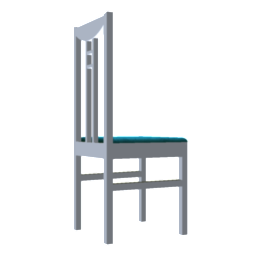}{2.0\figureheight}

  \figrow{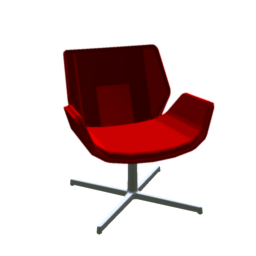}{3.0\figureheight}
  \figrow{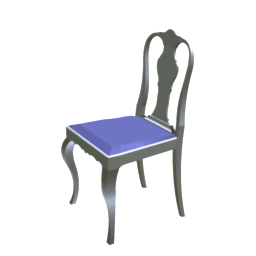}{4.0\figureheight}
  \figrow{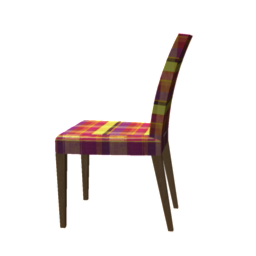}{5.0\figureheight}
  \figrow{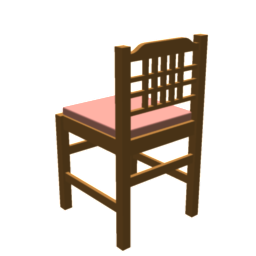}{6.0\figureheight}

  \end{tikzpicture}   
\end{subfigure}
 \hspace*{\fill}
\begin{subfigure}[b]{.49\textwidth} 

  \newcommand{\figg}[2]{\includegraphics[width={#1\figurewidth}]{./fig/car/#2}}

  \newcommand{\figrow}[3]{%
     \node [draw=white,thick,minimum width=\figurewidth,inner sep=0] at
       ({0*\figurewidth},#2) {\figg{#3}{#1_source}};%
     \node [draw=white,thick,minimum width=\figurewidth,inner sep=0] at
       ({1*\figurewidth},#2) {\figg{#3}{#1_gt}};%
     \node [draw=white,thick,minimum width=\figurewidth,inner sep=0] at
       ({2*\figurewidth},#2) {\figg{#3}{#1_mv3d}};%
     \node [draw=white,thick,minimum width=\figurewidth,inner sep=0] at
       ({3*\figurewidth},#2) {\figg{#3}{#1_sun}};%
     \node [draw=white,thick,minimum width=\figurewidth,inner sep=0] at
       ({4*\figurewidth},#2) {\figg{#3}{#1_chen}};%
        \node[draw=white,thick,minimum width=\figurewidth,inner sep=0] at
       ({5*\figurewidth},#2) {\figg{#3}{#1_ours}};
  }

  \begin{tikzpicture}

 \node[text width=.9\figurewidth,align=center,text centered,text depth = 0cm] at ({\figurewidth*0},{-.65*\figureheight}) {\footnotesize Input};
 \node[text width=.9\figurewidth,align=center,text centered,text depth = 0cm] at ({\figurewidth*1.1},{-.65*\figureheight}) {\footnotesize Target};
 \node[text width=.9\figurewidth,align=center,text centered,text depth = 0cm] at ({\figurewidth*2.1},{-.65*\figureheight}) {\footnotesize \cite{tatarchenko2016multi}};
 \node[text width=.9\figurewidth,align=center,text centered,text depth = 0cm] at ({\figurewidth*3.05},{-.65*\figureheight}) {\footnotesize \cite{sun2018multi}};
 \node[text width=.9\figurewidth,align=center,text centered,text depth = 0cm] at ({\figurewidth*4.1},{-.65*\figureheight}) {\footnotesize \cite{chen2019monocular}};
 \node[text width=.9\figurewidth,align=center,text centered,text depth = 0cm] at ({\figurewidth*5.07},{-.65*\figureheight}) {\footnotesize Ours};

  \figrow{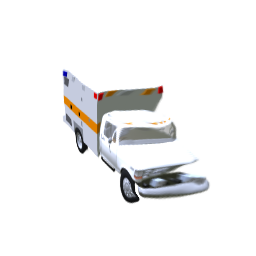}{0}{1.1}
  \figrow{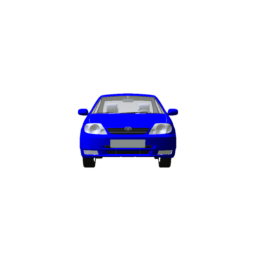}{1.0\figureheight}{1.23}
  \figrow{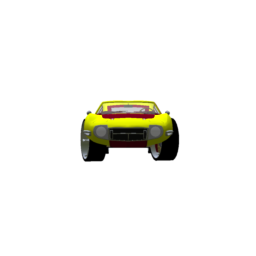}{2.0\figureheight}{1.2}
  \figrow{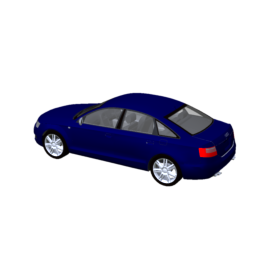}{3.0\figureheight}{1}
  \figrow{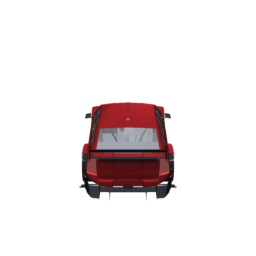}{4.0\figureheight}{1.2}
  \figrow{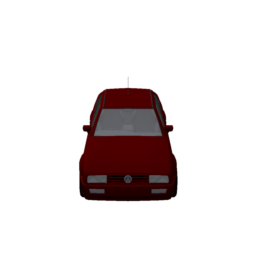}{5.0\figureheight}{1.1}

  \figrow{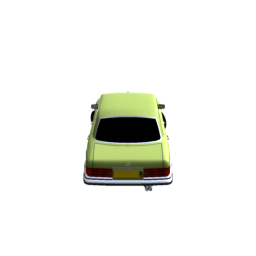}{6.0\figureheight}{1.1}
  
\node [draw=white,thick,minimum width=\figurewidth,inner sep=0] at
       ({0*\figurewidth},0) {\includegraphics[width=1.1\figurewidth]{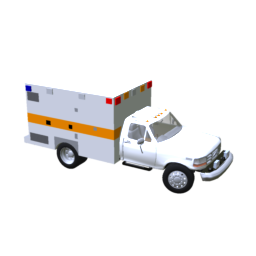}};%
\node [draw=white,thick,minimum width=\figurewidth,inner sep=0] at
       ({1*\figurewidth},0) {\includegraphics[width=1.1\figurewidth]{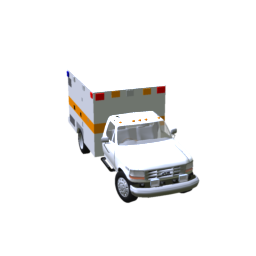}};%

  \end{tikzpicture}   
\end{subfigure}

  \caption{Results on ShapeNet objects. Our methods generate as structure-consistent predictions as pixel generation methods (for example, it can generate the missing chair legs compared to \cite{chen2019monocular}); on the other hand, our generated images are not blurry and include rich low-level details as image-based rendering methods (zoom in for better visualization).}
  \label{fig:shapenet}
\end{figure*}
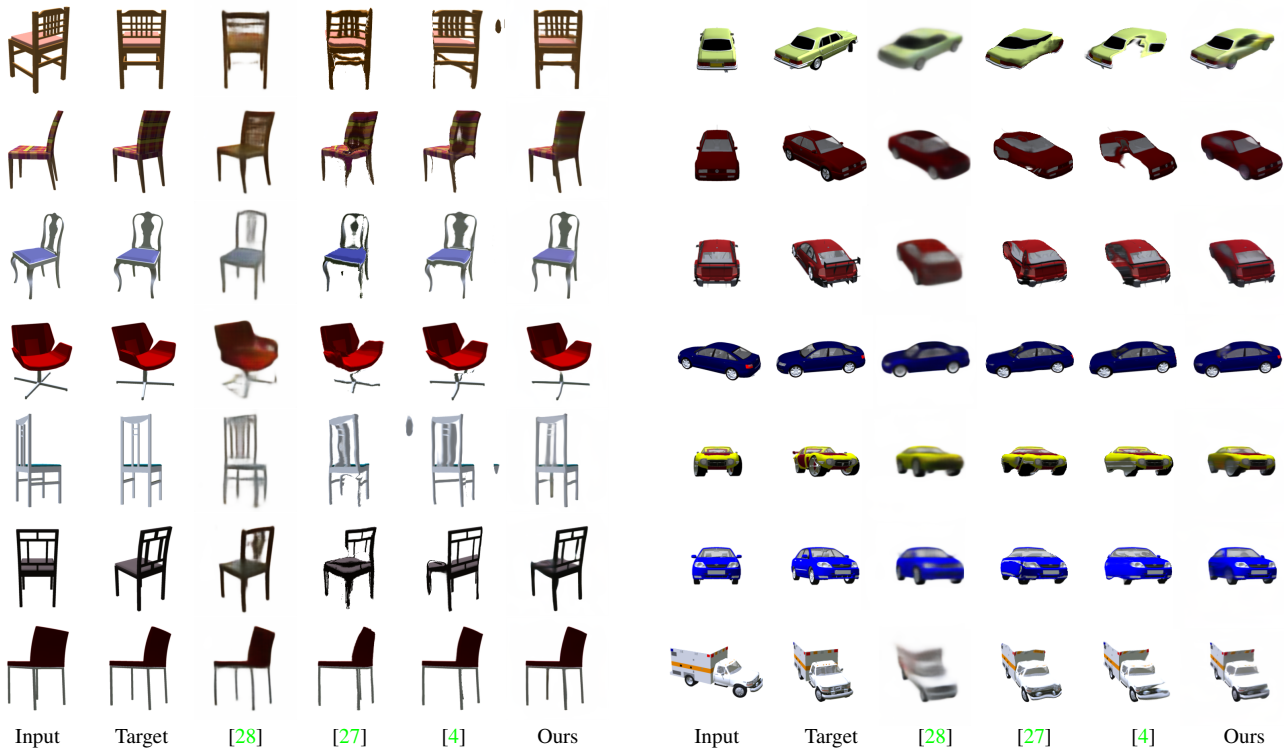

\begin{figure*}[!t]
  \centering
  \setlength{\figurewidth}{.166\textwidth}
  \setlength{\figureheight}{\figurewidth}

  \newcommand{\figg}[1]{\includegraphics[width=.98\figurewidth]{./fig/kitti/#1}}

  \newcommand{\figrow}[2]{%
     \node [draw=white,thick,minimum width=\figurewidth,inner sep=0] at
       ({0*\figurewidth},#2) {\figg{#1_source}};%
     \node [draw=white,thick,minimum width=\figurewidth,inner sep=0] at
       ({1*\figurewidth},#2) {\figg{#1_gt}};%
     \node [draw=white,thick,minimum width=\figurewidth,inner sep=0] at
       ({2*\figurewidth},#2) {\figg{#1_mv3d}};%
     \node [draw=white,thick,minimum width=\figurewidth,inner sep=0] at
       ({3*\figurewidth},#2) {\figg{#1_sun}};%
     \node [draw=white,thick,minimum width=\figurewidth,inner sep=0] at
       ({4*\figurewidth},#2) {\figg{#1_chen}};%
        \node[draw=white,thick,minimum width=\figurewidth,inner sep=0] at
       ({5*\figurewidth},#2) {\figg{#1_ours}};
  }

  \begin{tikzpicture}

 \node[text width=.9\figurewidth,align=center,text centered,text depth = 0cm] at ({\figurewidth*0},{-.65*\figureheight}) {\footnotesize Input};
 \node[text width=.9\figurewidth,align=center,text centered,text depth = 0cm] at ({\figurewidth*1},{-.65*\figureheight}) {\footnotesize Target};
 \node[text width=.9\figurewidth,align=center,text centered,text depth = 0cm] at ({\figurewidth*2},{-.65*\figureheight}) {\footnotesize \cite{tatarchenko2016multi}};
 \node[text width=.9\figurewidth,align=center,text centered,text depth = 0cm] at ({\figurewidth*3},{-.65*\figureheight}) {\footnotesize \cite{sun2018multi}};
 \node[text width=.9\figurewidth,align=center,text centered,text depth = 0cm] at ({\figurewidth*4},{-.65*\figureheight}) {\footnotesize \cite{chen2019monocular}};
 \node[text width=.9\figurewidth,align=center,text centered,text depth = 0cm] at ({\figurewidth*5},{-.65*\figureheight}) {\footnotesize Ours};

  \figrow{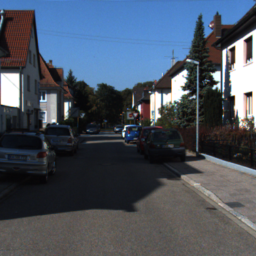}{0}  
  \figrow{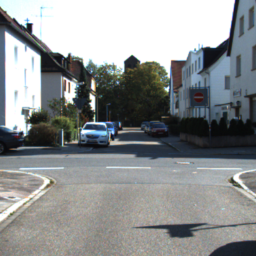}{1.0\figureheight}
  \figrow{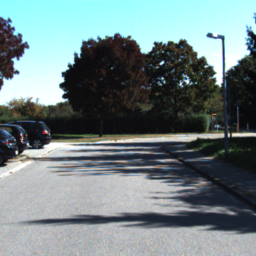}{2.0\figureheight}
  \figrow{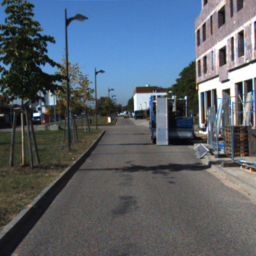}{3.0\figureheight}

  \end{tikzpicture}   
  \caption{Qualitative results on KITTI. Our method generate clear and structure consistent predictions, while pixel generation methods \cite{tatarchenko2016multi} struggle with blurry and image-based rendering methods \cite{chen2019monocular} suffer from distortion (see the house and truck on row~1, the street light pole on row~2, and the houses on row~3 and row~4).}
  \label{fig:kitti}
\end{figure*}

\subsection{Training Loss Functions}
The whole framework can be trained in an end-to-end manner since all modules in our pipeline are differentiable. For each input sample, only a single source image and the target image and their relative transformation are given. We optimize both the encoder, the depth decoder and the pixel decoder jointly. For pixel regression, we use multi-scale L1 reconstruction loss and VGG perceptual loss to encourage generating realistic images. To train the depth decoder in an unsupervised manner, we use the edge-aware smoothness loss and introduce a depth consistent loss to make more stable predictions.

\paragraph{Multi-scale Reconstruction Loss} To integrate learned prior knowledge and alleviate the negative impacts introduced by wrong depth prediction (\eg, distortion), we make multi-scale novel view predictions, finalizing the final prediction from coarse to fine-grained. The total reconstruction loss $\mathcal{L}_\textrm{reco}$ is the weighted combination of the individual losses at different scales in the pixel decoder:
\begin{equation}
\mathcal{L}_\textrm{reco}=  \sum_{i} w_i\,|\tilde{I}_\mathrm{t}^{i} - I_\mathrm{t}|,
\end{equation}
where $\tilde{I}_\mathrm{t}^{i}$ is the upsampled predicted target images and $w_i$ is the weight for results at different scale $i$. The weights decrease according to the resolution of prediction. Intuitively, compared to only considering the final prediction, the multi-scale reconstruction loss should help since it will produce gradients from larger receptive fields rather than small neighbourhoods. Also, as it needs to predict correct results at coarse levels, it can encourage the latent code to understand the 3D scene without the skip connections and alleviate the dependence of the skip connections, avoiding the distortion with inaccurate depth estimation.

\paragraph{VGG Perceptual Loss} Similar to \cite{park2017transformation}, besides L1 reconstruction loss, we adopt the VGG perceptual loss to get sharper synthesis results.  A pretrained VGG16 network is used for extracted features from generation results and ground-truth images, and the perceptual loss is the sum of feature distances (we use L1 distance) computed from a number of layers.

\paragraph{Depth Consistent Loss} To regularize the latent code and its depth prediction without supervision, we introduce a depth consistent loss. Intuitively, the depth decoder project the latent code into depth maps, so it should work for both extracted latent code $z$ and transformed latent code $\tilde{z}$. During the training, we also extract the latent code $z_t$ from the target view via the encoder. Instead of encouraging the transformed latent code $\tilde{z}_\mathrm{t}$ to be same with the latent code extracted target view $z_\mathrm{t}$, we encourage the distance of depth predictions to be small:
\begin{equation}
\mathcal{L}_\textrm{depth} = |\psi_d(z_\mathrm{t}) - \psi_d(\tilde{z}_\mathrm{t})|.
\end{equation}

\paragraph{Edge-aware Smoothness Loss} Similar to \cite{godard2017unsupervised}, we use an edge-aware smoothness loss that should encourage predicted depth maps to be locally smooth. The loss is weighted by an edge-aware term since the depth discontinuities often occur at image edges:
\begin{equation}
\mathcal{L}_\textrm{edge} = \frac{1}{N}\sum_{i,j}|\partial_x\tilde{D}_\mathrm{t}^{ij}|e^{-\| \partial_x I_\mathrm{t}^{ij} \|} + |\partial_y\tilde{D}_\mathrm{t}^{ij}|e^{-\| \partial_y I_\mathrm{t}^{ij} \|},
\end{equation}
where $\tilde{D}_\mathrm{t}$ is the predicted depth map of the target view and $I_\mathrm{t}$  is the ground-truth target view.

In conclusion, the final loss function for training the framework jointly will be 
\begin{equation}
\mathcal{L} = \lambda_m\mathcal{L}_\textrm{reco} + \lambda_v\mathcal{L}_\textrm{vgg} + \lambda_d\mathcal{L}_\textrm{depth} + \lambda_e\mathcal{L}_\textrm{edge},
\end{equation}
where the $\lambda_m$, $\lambda_v$, $\lambda_d$, and $\lambda_e$ are weights for different loss functions.

\begin{figure*}[!t]
  \centering
  \setlength{\figurewidth}{.1111\textwidth}
  \setlength{\figureheight}{\figurewidth}

  \newcommand{\figg}[1]{\includegraphics[width=.98\figurewidth]{./fig/ablation/#1}}

  \newcommand{\figrow}[2]{%
     \node [draw=white,thick,minimum width=\figurewidth,inner sep=0] at
       ({0*\figurewidth},#2) {\figg{#1_source}};%
     \node [draw=white,thick,minimum width=\figurewidth,inner sep=0] at
       ({1*\figurewidth},#2) {\figg{#1_gt}};%
     \node [draw=white,thick,minimum width=\figurewidth,inner sep=0] at
       ({2*\figurewidth},#2) {\figg{#1_ours}};%
     \node [draw=white,thick,minimum width=\figurewidth,inner sep=0] at
       ({3*\figurewidth},#2) {\figg{#1_wo_multi}};%
     \node [draw=white,thick,minimum width=\figurewidth,inner sep=0] at
       ({4*\figurewidth},#2) {\figg{#1_wo_vgg}};%
        \node[draw=white,thick,minimum width=\figurewidth,inner sep=0] at
       ({5*\figurewidth},#2) {\figg{#1_wo_edge}};
        \node[draw=white,thick,minimum width=\figurewidth,inner sep=0] at
       ({6*\figurewidth},#2) {\figg{#1_wo_conv0}};
        \node[draw=white,thick,minimum width=\figurewidth,inner sep=0] at
       ({7*\figurewidth},#2) {\figg{#1_wo_depth}};
        \node[draw=white,thick,minimum width=\figurewidth,inner sep=0] at
       ({8*\figurewidth},#2) {\figg{#1_wo_skip}};
  }

  \begin{tikzpicture}

\node[text width=.9\figurewidth,align=center,text centered,text depth = 0cm] at ({\figurewidth*0},{-.65*\figureheight}) {\footnotesize Input};
\node[text width=.9\figurewidth,align=center,text centered,text depth = 0cm] at ({\figurewidth*1},{-.65*\figureheight}) {\footnotesize Target};
\node[text width=.9\figurewidth,align=center,text centered,text depth = 0cm] at ({\figurewidth*2},{-.65*\figureheight}) {\footnotesize Ours};
\node[text width=.9\figurewidth,align=center,text centered,text depth = 0cm] at ({\figurewidth*3},{-.65*\figureheight}) {\footnotesize w/o multi};
\node[text width=.9\figurewidth,align=center,text centered,text depth = 0cm] at ({\figurewidth*4},{-.65*\figureheight}) {\footnotesize w/o $\mathcal{L}_\textrm{VGG}$};
\node[text width=.9\figurewidth,align=center,text centered,text depth = 0cm] at ({\figurewidth*5},{-.65*\figureheight}) {\footnotesize w/o $\mathcal{L}_\textrm{edge}$};
\node[text width=.9\figurewidth,align=center,text centered,text depth = 0cm] at ({\figurewidth*6},{-.65*\figureheight}) {\footnotesize w/o \textit{conv0}};
\node[text width=.9\figurewidth,align=center,text centered,text depth = 0cm] at ({\figurewidth*7},{-.65*\figureheight}) {\footnotesize w/o $\mathcal{L}_\textrm{depth}$};
\node[text width=.9\figurewidth,align=center,text centered,text depth = 0cm] at ({\figurewidth*8},{-.65*\figureheight}) {\footnotesize w/o skip};

  \figrow{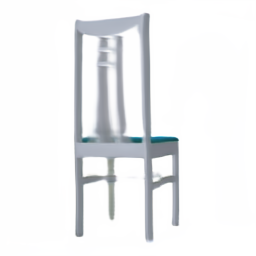}{0}  
  \figrow{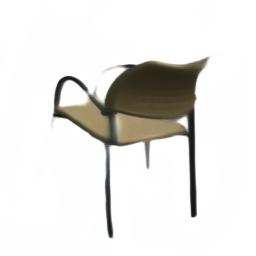}{1.0\figureheight}
  \figrow{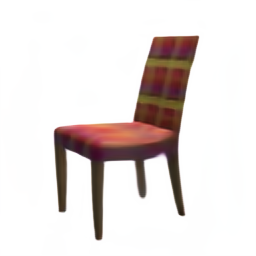}{2.0\figureheight}
  \figrow{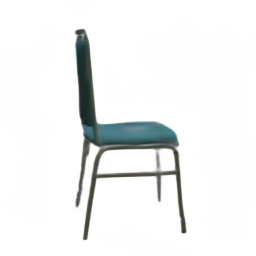}{3.0\figureheight}
        
  \end{tikzpicture}   

  \caption{Ablation study results. We compare the performance of our full model with its variants. Results show that the multi-scale loss help to generate thin structures (like the chair leg in the 1st and 3rd row of the 4th column). The $\mathcal{L}_\textrm{VGG}$ makes the results sharper. The skip connection from \textit{conv0} maintains the detailed features (like the pattern for 2nd row). The $\mathcal{L}_\textrm{depth}$ leads to more stable results. }
  \label{fig:ablation}
\end{figure*}

\section{Experiments}
To show that our method can combine the advantages of both image generation and 3D warping methods effectively, we compared our method with three state-of-the-art methods: one typical image generation method proposed by Tatarchenko~\etal~\cite{tatarchenko2016multi}, one image-based rendering method proposed by Chen~\etal~\cite{chen2019monocular}, which also share the similar TAE architecture with ours; and an explicit aggregation scheme proposed by Sun~\etal~\cite{sun2018multi} that predict confidence maps for both pixel generation results and flow prediction results. We replace the discrete one-hot viewpoint representation in \cite{tatarchenko2016multi} with cosine and sine values of the view angles. We jointly train the encoder, the depth decoder, and the pixel decoder using the Adam \cite{kingma2014adam} solver with $\beta_1= 0.9$ and $\beta_2 = 0.999$, and a learning rate of $10^{-4}$. For our methods, we add skip connection with feature maps \textit{conv0, conv2, conv3, conv4} from the encoder because of the best performance. To evaluate the predictions, we report numbers of mean absolute error L1 that measures per-pixel difference and the structural similarity (SSIM) index \cite{wang2004image} that indicates perceptual image quality. For L1 metric, smaller is better; for the SSIM metric, larger is better.

\subsection{Datasets}
We conduct our experiments on two different types of datasets: for objects we use ShapeNet synthetic dataset \cite{chang2015shapenet} and for real-world scenes we use KITTI Visual Odometry \cite{Geiger2012CVPR} dataset. More specifically, we select cars and chairs in the ShapeNet dataset. Generally, datasets with complicated structures and camera transformation will challenge the 3D understanding (\eg, depth estimation in our method), while datasets with rich textures will show if the methods can preserve the fine-grained details well. Among our selected datasets, the chairs have more complicated shapes and structures, but for each chair the texture is more simple. Reversely, the cars have much simpler shapes but there will be more colorful patterns on each car. For KITTI, the scene includes more objects, and translations are the main transformation between frames, unlike ShapeNet where rotation is the key transformation. In that case, the accurate depth estimation is less necessary (\cite{vangorp2013perception} shows that even approximating the scene as a single plane can give reasonable results), while the ability to recover the low-level details is more important for performance.

\paragraph{ShapeNet} ShapeNet is a large-scale synthetic dataset of clean 3D models \cite{chang2015shapenet}. For ShapeNet objects, we use rendered images with the dimension of $256\times256$ from 54 viewpoints (the azimuth from $0^{\circ}$ to $360^{\circ}$ with $20^{\circ}$ increments, and the elevation of $0^{\circ}$, $10^{\circ}$, and $20^{\circ}$) for each object. The training and test pairs are two views with the azimuth difference within the range $[-40^{\circ}, 40^{\circ}]$. For ShapeNet chairs, there are 558 chair objects in the training set and 140 chair objects in the test set; For ShapeNet cars, there are 5,997 car objects in the training set and 1,500 car objects in the test set.

\paragraph{KITTI} We use the KITTI odometry datasets since the ground-truth camera poses are provided. There are 11 sequences and each sequence contains around 2,000 frames on average. We restrict the training pairs to be separated by at most 7 frames.

For all selected datasets, we use the same train/test split as \cite{sun2018multi}. Since the evaluation samples in \cite{sun2018multi} are created for multi-view inputs, we randomly select the source views as our input view. In total, for ShapeNet chairs there are 42,834 pairs for testing; for ShapeNet cars there are 42,780 pairs for testing, and for KITTI there are 10,000 pairs for testing.

\subsection{ShapeNet Evaluation}
\cref{tbl:shapenet} shows the comparison results on ShapeNet objects. Our methods perform the best for both chair and car objects, showing that it can deal with complicated 3D structures of chairs as well as rich textures of cars at the same time. \cref{fig:shapenet} shows the qualitative results for all methods. The pixel generation method \cite{tatarchenko2016multi} produces blurry results and the object identity failed to be preserved because of the lack of low-level details. Image-based rendering methods \cite{chen2019monocular} suffer from the distortion. Our method combine the advantages of the two type of approaches. On the one hand, our method exploits learned prior knowledge and generate structure consistent predictions as \cite{tatarchenko2016multi} (\eg, our method can generate the missing chair leg in the last row and the missing tires of \cite{chen2019monocular} in row~5 and row~6); on the other hand, our generated images are not blurry and includes rich low-level details as \cite{chen2019monocular}. The aggregation method \cite{sun2018multi} is mainly designed for multi-view inputs, so their modules cannot benefit from the recurrent neural network architecture when the input is a single source image.

\begin{table}[]
\caption{Results on ShapeNet objects.  Our methods perform the best for both chair and car objects, showing that it can deal with complicated 3D structures of chairs as well as rich textures of cars.}
\label{tbl:shapenet}
\begin{tabular*}{\columnwidth}{@{\extracolsep{\fill}} l c c | c c }
\toprule
\multirow{2}{*}{\sc Methods} & \multicolumn{2}{c}{\sc Chair} & \multicolumn{2}{c}{\sc Car} \\
                         & \sc L1           & \sc SSIM        & \sc L1          & \sc SSIM       \\ \midrule
Tatarchenko~\cite{tatarchenko2016multi}              & 0.1043     & 0.8851     & 0.0491    & 0.9226   \\ 
Sun~\cite{sun2018multi}                      &      0.0810       &    0.8993        &       0.0444	      &     0.9282       \\ 
Chen~\cite{chen2019monocular}                 & 0.0769     & 0.9099    &      0.0396       &    0.9395        \\ 
Ours                     &     \bf{0.0584}      &   \bf{0.9256}          &    \bf{0.0286}         &   \bf{0.9493}        \\ \bottomrule
\end{tabular*}
\end{table}

\subsection{KITTI Evaluation}
We also evaluate all methods on KITTI to show that our model can capture low-level details well with the aid of skip connections. \cref{tbl:kitti} shows the quantitative results on the KITTI dataset. We achieve the best SSIM results and get comparable L1 performance as the aggregation method \cite{sun2018multi}. Since \cite{sun2018multi} uses adversarial loss for their pixel generation module, it can inpaint missing regions better than other methods. As a pixel generation method, our method is obviously better than \cite{tatarchenko2016multi} in terms of the fine-grained textures, which shows the effectiveness of aligned skip connections. Compared to image-based rendering method \cite{chen2019monocular}, we still perform better on both L1 error and SSIM.  In \cref{fig:kitti}, the qualitative results show the same finding. Generally, our method generates clear predictions, and preserves the structure better (check the house and truck in row 1, the street light pole in row 2, the houses in row 3 and row 4). 

\begin{table}[]
\centering
\caption{Results on KITTI. We achieve the best SSIM results, and the L1 performance is better than both the pixel generation method \cite{tatarchenko2016multi} and the image-based rendering method \cite{chen2019monocular}.}
\label{tbl:kitti}
\begin{tabular*}{\columnwidth}{@{\extracolsep{\fill}} l c c c}
\toprule
\multirow{2}{*}{\sc Methods\hspace*{6em}} & \multicolumn{2}{c}{\sc KITTI} & \hspace*{2em} \\
                         & \sc L1           & \sc SSIM            \\ \midrule
Tatarchenko~\cite{tatarchenko2016multi}              & 0.3119    & 0.6191      \\ 
Sun~\cite{sun2018multi}                      &      \bf{0.1868}      &    0.6582              \\ 
Chen~\cite{chen2019monocular}                 & 0.2354    & 0.6461          \\ 
Ours                     &     0.1985	 &   \bf{0.7043}                \\ \bottomrule
\end{tabular*}
\end{table}

\subsection{Ablation Studies}
To understand how different blocks of the framework play their roles, we conduct ablation studies on the ShapeNet chairs, since it is the most challenging selected dataset for 3D structures. \cref{tbl:ablation} and \cref{fig:ablation} show the performance of different variants. Firstly, we compare the performance of our baseline architecture and the image-based rendering method \cite{chen2019monocular}. In the baseline architecture, we use our two-decoders architecture and optimize with the simple L1 reconstruction loss only, while \cite{chen2019monocular} uses one decoder for depth estimation. Since our baseline architecture achieves better performance compared to \cite{chen2019monocular}, it shows the effectiveness of our framework on combining the warping methods and pixel generation methods. Moreover, we observe that without the skip connections, our method can be regarded as a typical pixel generation methods that still suffer the same issue as \cite{tatarchenko2016multi}, which also proves our assumption that the compact equivariant latent code cannot encode sufficient information for pixel regression and we need the assistance from skip connections.  Other numbers and qualitative results show the selected loss functions are both useful for boosting performance. The multi-scale reconstruction loss help to generate thin structures better (like the chair leg in the 1st and 3rd row of the 4th column).  The $\mathcal{L}_\textrm{VGG}$ makes the results sharper. The skip connection from \textit{conv0} maintains the detailed features (like the patterns for 2nd row). Both $\mathcal{L}_\textrm{edge}$ and $\mathcal{L}_\textrm{depth}$ regularize the depth prediction, especially the depth consistent loss $\mathcal{L}_\textrm{depth}$.The qualitative results show that without $\mathcal{L}_\textrm{depth}$ the synthesis target images will suffer from distortion because of the unstable quality of predicted depth maps.

\begin{table}[]
\centering
 \caption{Results for ablation studies. All designed modules and loss functions are both useful for boosting performance. The baseline arch means using our architecture with L1 reconstruction at the final resolution only, which shows that the two-decoders architecture helps already. }
 \label{tbl:ablation}

\begin{tabular*}{\columnwidth}{ l c c}
\toprule
\sc Methods \hspace*{9em}             & \sc L1     & \sc SSIM   \\ \midrule
Chen~\cite{chen2019monocular}                   & 0.0769 & 0.9099 \\ 
Our baseline architecture    & 0.0611 & 0.9228 \\ 
Our final arch       & \bf{0.0584} & \bf{0.9256} \\ 
~~~w/o skip connections &   0.0949	   &     0.8971 \\ 
~~~w/o \textit{conv0}            & 0.0654 & 0.9166 \\ 
~~~w/o multi-scale loss & 0.0602 & 0.9231 \\ 
~~~w/o $\mathcal{L}_\textrm{VGG}$                   & 0.0603 & 0.9236 \\ 
~~~w/o $\mathcal{L}_\textrm{depth}$                    & 0.1109 & 0.8862 \\ 
~~~w/o $\mathcal{L}_\textrm{edge}$                   & 0.0618 & 0.9212 \\ \bottomrule
\end{tabular*}
\end{table}

\subsection{Depth estimation}
\begin{figure}[!t]
  \setlength{\figurewidth}{.24\columnwidth}
  \setlength{\figureheight}{\figurewidth}
\begin{subfigure}[b]{.22\columnwidth} 
  \begin{tikzpicture}
  \node [draw=white,thick,minimum width=\figurewidth,inner sep=0] at (0,0) 
{\includegraphics[width=1.12\figurewidth]{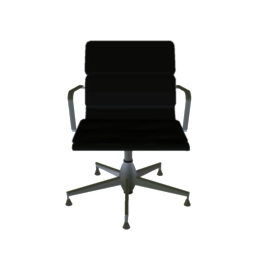}};%

     \node [draw=white,thick,minimum width=\figurewidth,inner sep=0] at (0,1.0\figureheight) 
{\includegraphics[width=1.12\figurewidth]{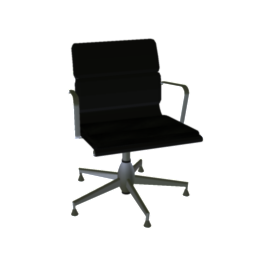}};%

\node[text width=.9\figurewidth,align=center,text centered,text depth = 0cm] at ({\figurewidth*0},{-.65*\figureheight}) {\footnotesize Target};

\node[text width=.9\figurewidth,align=center,text centered,text depth = 0cm] at ({\figurewidth*0},{1.65*\figureheight}) {\footnotesize Source};

  \end{tikzpicture}   
\end{subfigure}
\hspace*{\fill}
\begin{subfigure}[b]{.65\columnwidth} 
  \begin{tikzpicture}
  \node [draw=white,thick,minimum width=\figurewidth,inner sep=0] at (0,0) 
{\includegraphics[width=2.8\figurewidth]{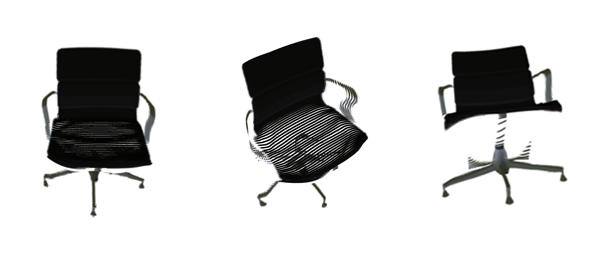}};%

     \node [draw=white,thick,minimum width=\figurewidth,inner sep=0] at (0,1.0\figureheight) 
{\includegraphics[width=2.8\figurewidth]{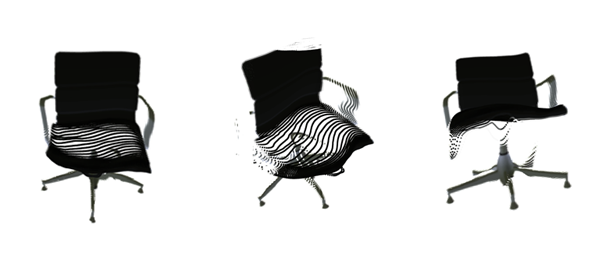}};%

\node[text width=.9\figurewidth,align=center,text centered,text depth = 0cm] at ({\figurewidth*0},{-.65*\figureheight}) {\footnotesize Ours};

\node[text width=.9\figurewidth,align=center,text centered,text depth = 0cm] at ({\figurewidth*0},{1.65*\figureheight}) {\footnotesize Chen~\cite{chen2019monocular}};
  
  \end{tikzpicture}   
\end{subfigure}

  \caption{Unsupervised depth prediction results that are visualized as point clouds depicted from different viewing angles. The left col shows the source input and the target view. The right cols show the comparison of the predicted depth map of the target view (The upper row shows results from \cite{chen2019monocular} and the bottom row shows our results). }
  \label{fig:depth}
\end{figure}
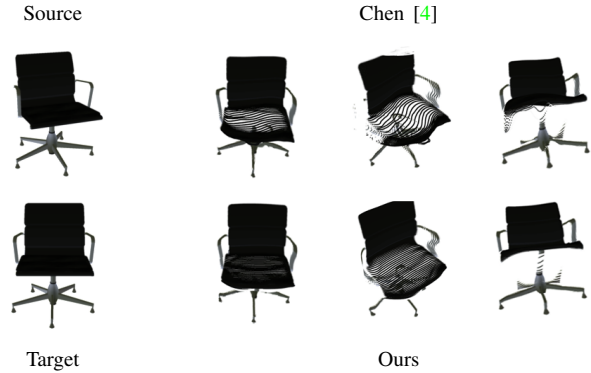

For the selected datasets we used for evaluation, since the accuracy of the predicted depth maps affect the most on ShapeNet chairs, we evaluate four depth metrics on ShapeNet chairs. L1-all compute the mean absolute difference, L1-rel compute the mean absolute relative difference $\textrm{L1-rel} = \frac{1}{n}\sum_i|\textrm{gt}_i-\textrm{pred}_i|/\textrm{gt}_i$, and L1-inv metric is mean absolute difference in inverse depth $\textrm{L1-inv}=\frac{1}{n}\sum_i|\textrm{gt}_i^{-1} - \textrm{pred}_i^{-1}|$. Except L1 metrics, we also utilize $\textrm{sc-inv}=(\frac{1}{n}\sum z_i^2 - \frac{1}{n^2}(\sum z_i)^2)^{\frac{1}{2}}$, where $z_i=\log(\textrm{pred}_i) - \log(\textrm{gt}_i)$ . L1-rel normalizes the error, L1-inv puts more importance to close-range depth values, and sc-inv metric is scale-invariant. \cref{tbl:depth} shows that our predicted depth is more accurate compared to \cite{chen2019monocular}, which also explains why we can achieve better results than their method. \cref{fig:depth} also visualized the predicted map as point clouds from different viewing angles, which shows that our predicted depth map is less distorted.

\begin{table}[]
  \centering
 \caption{Depth estimation results on ShapeNet chairs. L1-all compute the mean absolute difference,  L1-rel normalizes the error, L1-inv puts more importance to close-range depth values, and sc-inv metric is scale-invariant.}
 \label{tbl:depth}
\begin{tabular*}{\columnwidth}{@{\extracolsep{\fill}} l c c c c}
\toprule
     & \sc L1-all & \sc L1-rel & \sc L1-inv & \sc sc-inv \\ \midrule
Chen~\cite{chen2019monocular}  & 0.0707 & 0.0360 & 0.0189 & 0.0583 \\ 
Ours & \bf{0.0610} & \bf{0.0305} & \bf{0.0161} & \bf{0.0523} \\
\bottomrule
\end{tabular*}
\end{table}

\section{Conclusion and Discussion}
In this paper, we propose an image generation pipeline that can take advantage of explicit correspondences. We predict the depth map of the target view from a single input view and warp the feature maps of the input view. The warping enables the skip connections to transfer low-level details, so our method can produce clear predictions. Experiment results show that our methods perform better than warping methods and pixel generation methods, alleviating distortion and blurry issues.

Currently for depth prediction, the TAE architecture can only provide a coarse depth map without skip connections, which cannot get correct predictions for thin structures like the arm of chairs. Investigating how to obtain accurate estimation for thin structures can be the future work to further improve the performance.

The code for the experiments can be found at \url{https://github.com/AaltoVision/warped-skipconnection-nvs}.

\paragraph{Acknowledgements} We acknowledge the computational resources provided by the Aalto Science-IT project. This researchwas supported by the Academy of Finland grants 324345 and 309902.

{\small
\bibliographystyle{ieee_fullname}
\bibliography{egbib}
}

\end{document}